\documentclass{article}
\usepackage{neurips_2024}

\usepackage{graphicx}
\usepackage[utf8]{inputenc} % allow utf-8 input
\usepackage[T1]{fontenc}    % use 8-bit T1 fonts
\usepackage{hyperref}       % hyperlinks
\usepackage{url}            % simple URL typesetting
\usepackage{booktabs}       % professional-quality tables
\usepackage{amsfonts}       % blackboard math symbols
\usepackage{nicefrac}       % compact symbols for 1/2, etc.
\usepackage{microtype}      % microtypography
\usepackage{xcolor}         % colors
\usepackage{svg}
\usepackage{amsmath}
\usepackage{multirow}
\usepackage{array}
\usepackage{geometry}
\usepackage{colortbl}
\usepackage{quoting}
\usepackage{subcaption}
\quotingsetup{leftmargin=1cm}
\usepackage{enumitem}

\newcommand{\tab}[1]{Table~\ref{#1}}
\newcommand{\fig}[1]{Fig.~\ref{#1}}
\newcommand{\sect}[1]{\S\ref{#1}}

\newcommand{\eg}{\textit{e.g.}, } 
\newcommand{\ie}{\textit{i.e.}, }

\makeatletter
\renewcommand\@makefnmark{\textsuperscript{\dagger}}
\renewcommand\@fnsymbol[1]{\dagger}
\makeatother

\title{The Death of Schema Linking? Text-to-SQL in the Age of Well-Reasoned Language Models}

\author{
Karime Maamari$^{1}$ \quad  Fadhil Abubaker$^{1}$ \quad Daniel Jaroslawicz$^{1}$ \quad Amine Mhedhbi$^{2}$\\
$^1$Distyl AI \quad $^2$Polytechnique Montreal\\
\texttt{\{karime,fadhil,daniel\}@distyl.ai}\\
\texttt{amine.mhedhbi@polymtl.ca}
}

\begin{document}
\maketitle

\begin{abstract}
Schema linking is a crucial step in Text-to-SQL pipelines. 
Its goal is to retrieve the relevant tables and columns of a target database for a user's query while disregarding irrelevant ones. 
However, imperfect schema linking can often exclude required columns needed for accurate query generation. 
In this work, we revisit schema linking when using the latest generation of large language models (LLMs). 
We find empirically that newer models are adept at utilizing relevant schema elements during generation even in the presence of large numbers of irrelevant ones. 
As such, our Text-to-SQL pipeline entirely forgoes schema linking in cases where the schema fits within the model's context window in order to minimize issues due to filtering required schema elements. 
Furthermore, instead of filtering contextual information, we highlight techniques such as augmentation, selection, and correction, and adopt them to improve the accuracy of our Text-to-SQL pipeline.
Our approach ranks first on the BIRD benchmark achieving an accuracy of 71.83\%.
\end{abstract}

\section{Introduction}

We address the task of Text-to-SQL: generating a database-executable SQL query given a natural language inquiry \citep{androutsopoulos1995naturallanguageinterfacesdatabases,DBLP:journals/ftdb/QuamarELO22}.Text-to-SQL is crucial in democratizing data access as it allows querying databases using natural language The advent of large language models (LLMs) has significantly advanced Text-to-SQL by simplifying the translation of natural language into SQL.

\begin{figure}[t]
  \centering
\centerline{\includegraphics[width=1.\columnwidth]{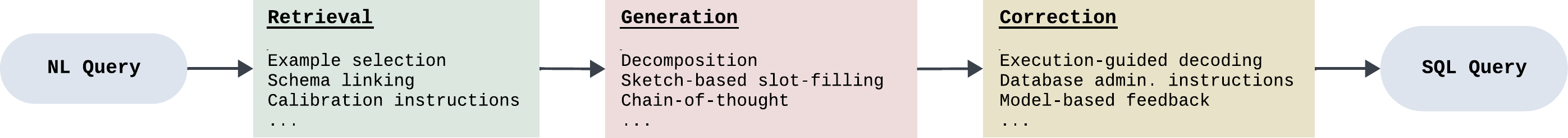}}
  \caption{A typical Text-to-SQL pipeline comparised of retrieval, generation and correction stages.}
  \label{fig:pipeline}
\end{figure}

LLM-based Text-to-SQL approaches typically follow a multi-stage generation pipeline, as shown in Fig. \ref{fig:pipeline} \citep{hong2024nextgenerationdatabaseinterfacessurvey,Li2024TheDO, liu2024surveynl2sqllargelanguage, zhang2024naturallanguageinterfacestabular}. 
The pipeline begins with a retrieval stage to collect contextual knowledge 
such as the definition of terms and database schema elements. 
This is followed by a generation stage, where an LLM produces a candidate SQL query. 
Finally, the correction stage regenerates the SQL as needed based on encountered errors.

Selecting relevant elements of the database schema (tables and columns) -- known as \emph{schema linking} -- provides the necessary context for the LLM to produce correct SQL in the downstream generation stage. Effective schema linking implies retrieving \emph{all} the relevant database components associated with the natural language query. Missing even a single required column results in incorrect executable SQL. 
Thus, it is vital to ensure that all essential columns are retrieved. 
However, this does not mean that it should be overly inclusive. 
Research has shown that false positives, \ie the number of irrelevant columns passed to the LLM, can often degrade text-to-SQL accuracy. 
For example, \cite{Floratou2024NL2SQLIA} showed that even when the entire database schema fits into the context of an LLM, it is still advantageous to perform schema linking. 
At the same time, attempts to prune irrelevant columns may also remove some required ones. 
Thus, schema linking traditionally contains an inherent trade-off between minimizing false positives while also preserving relevant context. 

As LLM reasoning capabilities improve, we challenge the conventional wisdom that schema linking is necessary for accurate Text-to-SQL when the schema fits within the model's context window. 
We find empirically that as model reasoning improves, the benefits of reducing false positives diminishes, \ie newer models are more capable of sifting through the schema to identify relevant columns compared to older models \citep{laban2024summaryhaystackchallengelongcontext, li2024needlebenchllmsretrievalreasoning}. 
This effect might be similar to how the latest LLMs when probed can reliably recall `a needle in multiple millions of
tokens of ''haystack``'~\citep{needle-hay,DBLP:journals/corr/abs-2403-05530}. 
For these models, schema linking is unnecessary and can even be detrimental, as it may filter out essential columns.
Instead, we present alternatives to schema linking that improve accuracy without schema information loss. 
Our approach based on these insights currently ranks first in accuracy at 71.83\% on the BIRD benchmark \citep{li2023llmservedatabaseinterface}. 

\section{Preliminaries}
We first outline the key elements of the Text-to-SQL pipeline, with a focus on schema linking and its implications.

\subsection{Text-to-SQL Pipeline}
In current state-of-the-art approaches, Text-to-SQL uses multi-sage pipelines comprised of retrieval, generation, and correction stages. 

The retrieval stage gathers relevant contextual information, including schema elements, domain knowledge, and example queries \citep{dong2023c3, gao2023texttosqlempoweredlargelanguage, lee2024mcssql, pourreza2023dinsql, wang2024macsql}. 

The generation stage often involves more than just producing a candidate SQL query associated given an input context. 
Rather, approaches frequently augment the generation process through techniques like decomposed generation \citep{maamari2024endtoendtexttosqlgenerationanalytics, pourreza2023dinsql, wang2024macsql} and chain-of-thought prompting \citep{wei2023chainofthought}. In addition, most approaches employ methods like self-consistency and multi-choice selection to produce multiple results, selecting the best outcome \citep{dong2023c3, gao2023texttosqlempoweredlargelanguage, lee2024mcssql}. 

The retrieval and generation stages often contain various techniques chained together. These techniques can be broadly categorized as filtering or augmenting, \ie techniques can either strip away unnecessary contextual information or attempt to provide additional useful context. 

Finally, the correction stage will often employ some combination of execution-based feedback \citep{wang2018robusttexttosqlgenerationexecutionguided, lin2020bridgingtextualtabulardata, he2019xsqlreinforceschemarepresentation, lyu2020hybridrankingnetworktexttosql} or model-based feedback \citep{talaei2024chess, askari2024magicgeneratingselfcorrectionguideline, wang2018robust} to correct the generated query. 

\subsection{Schema Linking}
Within the retrieval stage, schema linking leverages sophisticated prompting techniques to produce variable-length representations, hierarchically retrieve components, and iteratively process the schema \citep{talaei2024chess, dong2023c3, pourreza2023dinsql, lee2024mcssql, wang2024macsql, gao2023texttosqlempoweredlargelanguage}.  

These techniques can vary in i) how they represent the schema and ii) how they perform linking on that representation. For instance, some may represent the schema in natural language, while others utilize code-like structures \citep{gao2023texttosqlempoweredlargelanguage}. The approach to linking can also differ across techniques, with some directly filtering the schema \citep{dong2023c3, lee2024mcssql, talaei2024chess}, and others using intermediate representations to identify relevant tables and columns \citep{qu2024generationalignitnovel}. The choice of schema representation and linking strategy can have a significant influence on accuracy \citep{gao2023texttosqlempoweredlargelanguage}. This variability underscores the importance of selecting an appropriate method tailored to the specific requirements of the task, as the degree of filtering can directly impact the loss of schema information incurred during the schema linking process.

Most prior investigations of schema linking regardless of approach have arrived at the same conclusion -- schema linking yields meaningful gains in accuracy \citep{guo2019complex, Li2024TheDO, talaei2024chess}. However, these explorations used LLMs that are more sensitive to the presence of irrelevant columns (false positives) as contextual information, where reducing the false positives yielded meaningful gains in performance \citep{Floratou2024NL2SQLIA}. 

\section{Experimental Setup}

Our experimental analysis guides the design and implementation of our proposed Text-to-SQL pipeline~(\sect{sec:disc}). Our experiments aim to answer empirically three research questions (RQs):
\begin{itemize}[leftmargin=4em]
	\setlength{\itemsep}{2pt}
	\setlength{\parskip}{0pt}
	\setlength{\parsep}{0pt} 
	\item[\emph{RQ1.}] How does the inclusion of irrelevant schema elements impact SQL generation?
    \item[\emph{RQ2.}] How can the trade-off between precision and recall in schema linking techniques be characterized, and what is its downstream impact on SQL generation? 
    \item[\emph{RQ3.}] How do other techniques and stages within Text-to-SQL pipelines, aside from schema linking, comparatively impact SQL generation?
\end{itemize}

Next, we cover the details of our setup (datasets and models), and our methodology and metrics.

\subsection{Datasets}

We conducted our experiments using the BIRD dataset~\citep{li2023llmservedatabaseinterface}, 
which is widely considered to be the most challenging Text-to-SQL benchmark. % available at time of submission. 
BIRD contains queries from 95 databases spanning a wide breadth of domains, such as education and hockey, and is designed to mimic the complexity of real-world databases. 
This complexity arises from its \emph{``dirty''} format, where data, queries, and external knowledge may contain flaws --- queries can be incorrect, database columns might be improperly described, and databases can contain null values and unexpected encodings. 
Our evaluation set consisted of $10\%$ of the entries from each database in the dev set, as done in evaluations in prior work~\citep{talaei2024chess}. 
Our training set consisted of $500$ of the $9,428$ available samples in the BIRD training dataset.

\subsection{Models}
\label{subsec:models}

We used the following language models with context windows sufficiently large to accommodate the entire schema for each query in the BIRD evaluation set:

\begin{tabular}{p{0.001\textwidth}
p{0.33\textwidth} p{0.28\textwidth} p{0.31\textwidth}}
& 
\texttt{ft:GPT-4o (fine-tuned)}

\texttt{GPT-4o}

\texttt{GPT-4o-Mini}

\texttt{GPT-4-Turbo}

&

\texttt{Llama 3.1-405b}

\texttt{Llama 3.1-70b}

\texttt{Llama 3.1-8b}

\texttt{Deepseek Coder-V2}

&

\texttt{Claude 3.5 Sonnet}

\texttt{Claude 3 Opus}

\texttt{Mixtral-8x22B}

\texttt{Gemini 1.5 Pro}

\end{tabular}

\vspace{-1em}
We overview the fine-tuning approach of \texttt{GPT-4o} momentarily under methodology~(\sect{sec:methodology}). 

\subsection{Methodology}
\label{sec:methodology}

We design an empirical experiment for each research question:
\begin{itemize}[leftmargin=4.25em]
	\setlength{\itemsep}{2pt}
	\setlength{\parskip}{0pt}
	\setlength{\parsep}{0pt} 
	\item[\emph{Exp 1.}] We use a simplified Text-to-SQL pipeline consisting of only schema linking within the retrieval stage followed by a single attempt at generation. 
    For each query, we provide all the required columns and vary the amount of irrelevant columns to look at the impact of irrelevant columns retrieved on the generation. 
    \item[\emph{Exp 2.}] Using the same simplified pipeline, we introduce different implementations of schema linking that vary in precision and recall and analyze their impact on generation. 
    \item[\emph{Exp 3.}] We introduce augmentation, selection, and correction techniques on top of the simplified pipeline without and with schema linking. We run an ablation study to understand the relative impact of each technique on end-to-end accuracy.
\end{itemize}

\emph{Runs and input/output structure}. In all runs, the temperature was set to zero and structured output was used whenever possible. 
Given that not all models are capable of reliable structured output generation, 
the generated SQL query was fed through an identity call by \texttt{GPT-4o-Mini} in \texttt{JSON} mode to handle any potential issues with output formatting. 
The relative position of any schema element in an input prompt follows the same ordering as that provided by the schema definition of the benchmark. 

\emph{Fine-tuning \texttt{GPT-4o}}. Fine-tuning is done iteratively. 
At each iteration, we first fine-tune on a sample of $N$ triples: natural language query, SQL query, and schema elements. 
For each query, the schema includes all required columns and a random number of irrelevant columns picked uniformly at random. 
We then evaluate on BIRD's dev set. 
For each failed query, we prompt the model to reason about the failure, aggregate the reasoning across queries, and use it to pick the sample for the next iteration.  
Finally, based on the reasoning, we select a new sample of size $N$. 
The iterations stop once a pre-determined accuracy score is reached. 

\begin{figure}[t]
  \centering
\centerline{\includegraphics[width=1.\columnwidth]{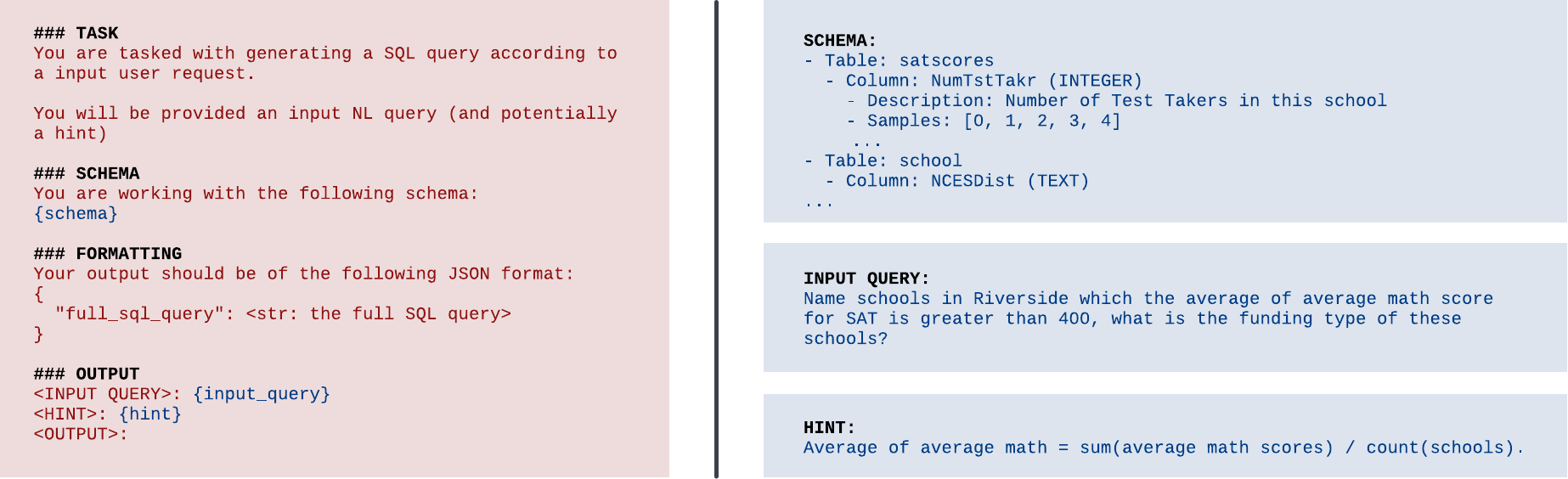}}
  \caption{\textbf{(Red}) Structure of SQL Generation prompt given input query, hint, and schema; \\
  \textbf{(Blue}) Examples of a schema, input query, and query hint which act as contextual inputs to a prompt.}
  \label{fig:prompts-gen}
\end{figure}

\emph{Generation prompts}. \fig{fig:prompts-gen} shows the structure of the prompt used for SQL generation as well as an example schema, input query, and query hint. 

\subsection{Metrics}

We rely on three different metrics in our experimental analysis:
\begin{itemize}[leftmargin=2em]
	\setlength{\itemsep}{2pt}
	\setlength{\parskip}{0pt}
	\setlength{\parsep}{0pt} 
	\item[-] \emph{Execution Accuracy} (EX): The metric used by the BIRD benchmark to evaluate end-to-end Text-to-SQL pipelines. 
    It is the proportion of queries for which the output of the predicted SQL query is identical to that of the ground truth SQL query. 
    We report EX as a percentage over queries in the evaluation set. 
    \item[-] \emph{False Positive Rate} (FPR): %  
    For a given query, the proportion of irrelevant schema columns retrieved over the total number of retrieved columns. 
    We report its average over queries in the evaluation set. 
    \item[-] \emph{Schema Linking Recall} (SLR): 
    The proportion of queries for which all required columns are retrieved over the total number of queries. 
    We use SLR as the downstream generation requires all required columns to be retrieved to be correct. 
\end{itemize}

All queries in evaluation sets are across multiple databases. 
Note that BIRD also has a second metric: valid efficiency score. 
It assesses the efficiency of correctly predicted queries by comparing their execution speed to those of the corresponding ground truth queries. 
In our experiments, we focus on EX, as our research questions are primarily concerned with SQL generation accuracy. 

\section{Results}

\subsection{Experiment 1: Impact of False Positives on Accuracy Given Perfect SLR}
\label{exp1}

\begin{figure*}[t]
  \begin{minipage}[t]{0.686\columnwidth}
    \centering
    \includegraphics[width=1\columnwidth]{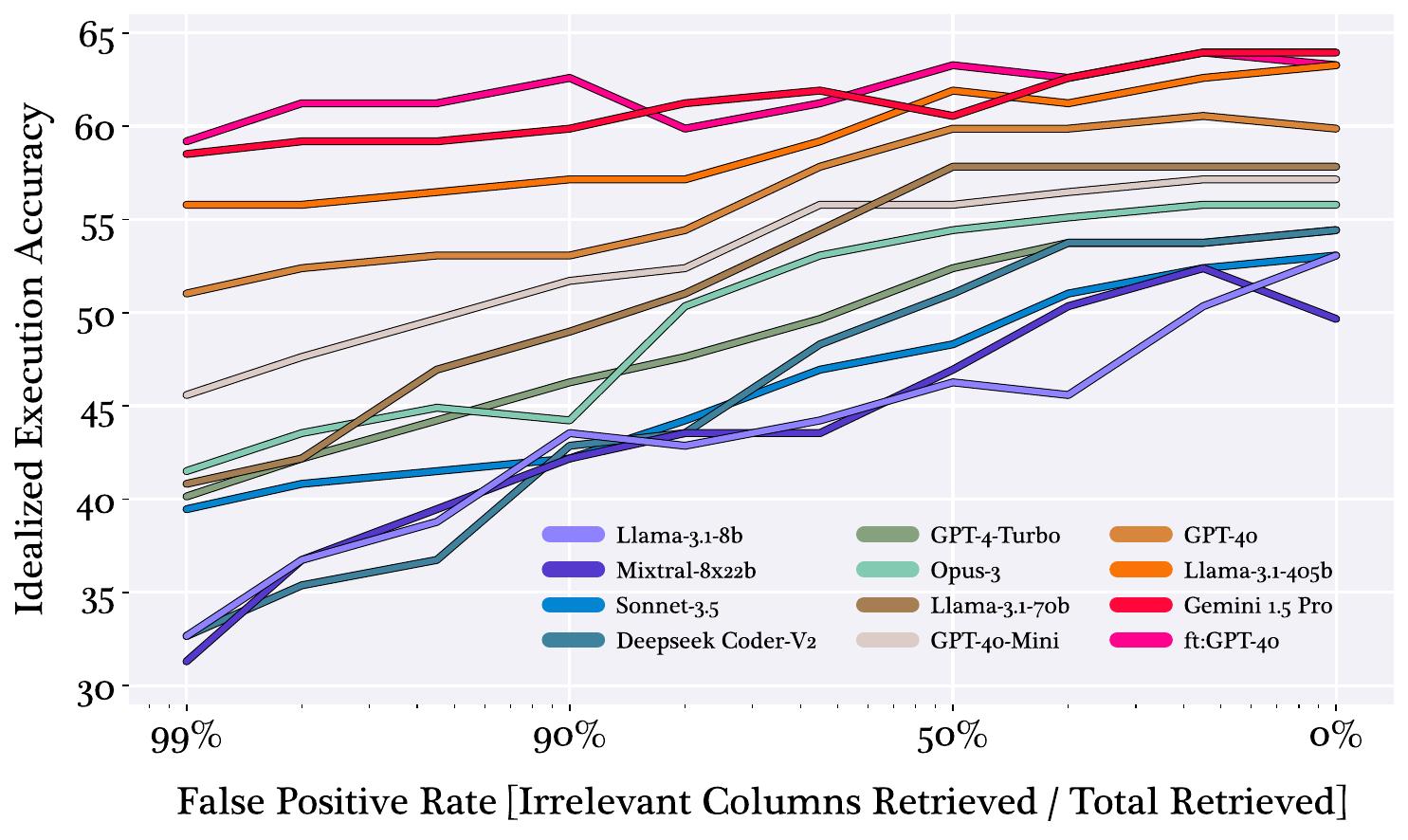}
    \subcaption{The idealized execution accuracy (IEX) as the false positive rate (FPR) varies from 99\% to 0\%.}
    \label{fig:2}
  \end{minipage}
  \hspace{0.5em}
  \begin{minipage}[t]{0.274\columnwidth}
    \centering
    \includegraphics[width=0.987\columnwidth]{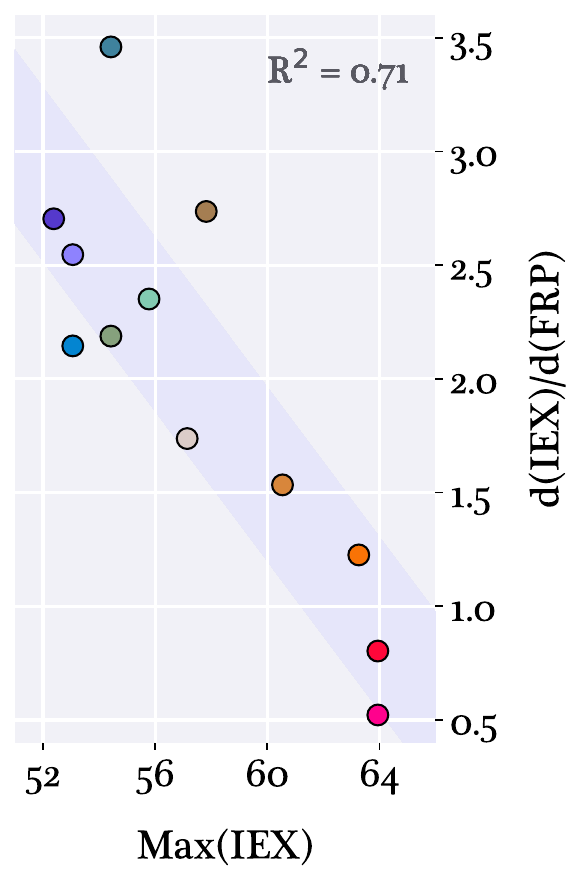}
    \subcaption{Capability and sensitivity relationship.}
    \label{fig:3}
  \end{minipage}
  \caption{Idealized execution accuracy (IEX), \ie with perfect schema linking recall (SLR), as the false positive rate (FPR) varies for different LMs and the relationship between their capability (maximum IEX) and sensitivity to false positives.}
\end{figure*}

In this first experiment, we assess the impact of retrieving irrelevant columns on SQL generation accuracy. 
We create a scenario with perfect schema linking recall such that SQL generation issues are not due to missing required columns. 

To implement the experiment, we mock the 
schema linker as follows. 
We build an oracle that identifies all columns necessary for a given query using \texttt{SQLGLOT} (SQL Parser and Transpiler). 
For an input query, the mocked schema linker uses the oracle to retrieve all required columns and injects a pre-defined rate of false positives (irrelevant columns retrieved / total columns retrieved). 
To inject it, the mocked schema linker samples the irrelevant columns uniformly at random from the target database. 
If there is not a sufficient number of columns in the target database, 
\eg there are $10$ required columns and $900$ irrelevant columns are needed to produce a $99\%$ false positive rate, yet only $50$ columns exist in the target database, 
it supplements from other databases with columns not conflicting in name. 
We use a simple pipeline: a retrieval stage containing only the mocked schema linker followed by a zero-shot generation stage.  

We run the pipeline using the $12$ selected models (\sect{subsec:models}) for generation
while applying an equal false positive rate to each query in the evaluation set. 
We run for $10$ different rates that are equally log-spaced from $99\%$ and $0\%$. 
We refer to the execution accuracy under perfect schema linking recall as the idealized execution accuracy (IEX).
We track the change in IEX as the false positive rate changes.  
The results in \fig{fig:2} 
show the broad trend that IEX improves as the rate of false positives decreases;  
specifically, the stage of SQL generation improves as less irrelevant columns are included as contextual information. 
At one extreme with a maximum rate of false positives ($99\%$), 
there is a $\sim28\%$ IEX difference between the worst and best performing models. 
In the other extreme with no false positives ($0\%$), the IEX difference between the worst and best performing models gets reduced to $\sim14\%$. 

We use a model's maximum IEX in this experiment as a proxy for its SQL generation capability. 
Models with higher generation capability, such as \texttt{Gemini 1.5 Pro}, demonstrate greater resilience to false positives compared to lower-performing ones like \texttt{Llama 3.1-8b}. 
We characterize a model's resilience to false positives by the relative change in its SQL generation capability as the false positive rate changes. 
We define a metric for sensitivity to false positives as the proportion of change in IEX over the proportion of change in the false positive rate (FPR): $d(IEX)/d(FRP)$. 
As such, sensitivity to false positives is the slope derived from the model’s (IEX, FPR) data points. 
\fig{fig:3} depicts a strong negative correlation between a model's SQL generation capability (maximum IEX) and its sensitivity to false positives. 

\emph{Empirical Observation:} As the model’s SQL generation capability improves, its sensitivity to the presence of irrelevant columns as contextual information for generation decreases. 
Perhaps surprisingly, both of these capabilities go hand-in-hand where models that are generally better at SQL generation are also more resilient to large amounts of irrelevant context. 

\subsection{Experiment 2: Impact of False Positives on Accuracy Given Actual SLR} 

In Experiment 1, all necessary columns for generation are provided, regardless of the false positive rate.
However, in practice, decreasing the amount of false positives requires pruning, which carries the risk of excluding required columns. 
Here we assess the extent to which schema linking affects recall of required columns and the downstream impact of imperfect recall on generation. 

\begin{figure}[t]
  \centering
\centerline{\includegraphics[width=1.\columnwidth]{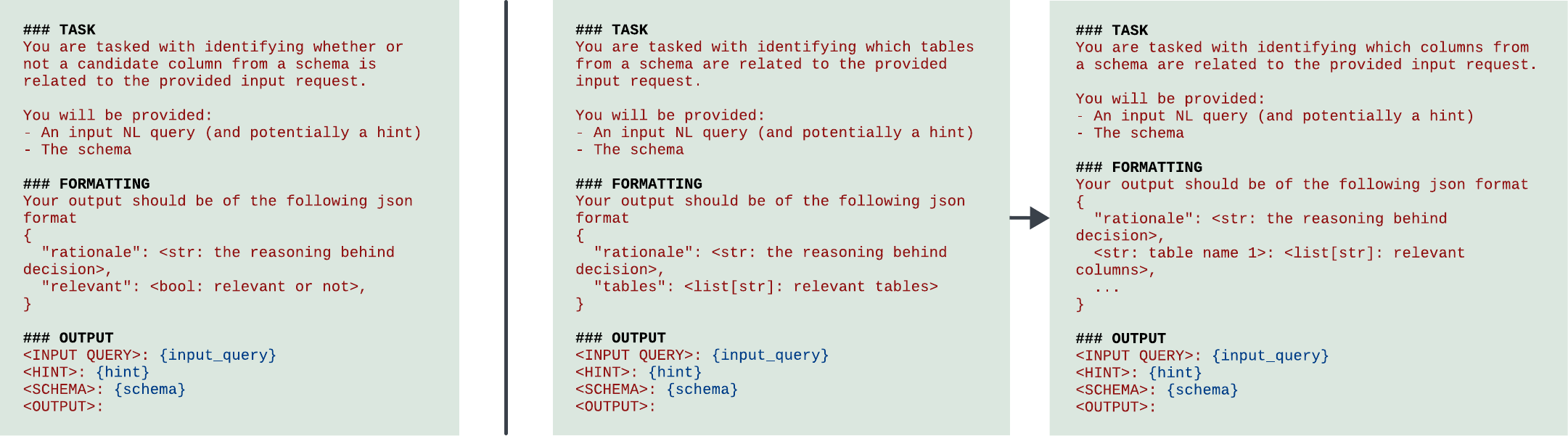}}
  \caption{Prompts used for schema linking. 
  \textbf{(Left}) Single-Column Schema Linking (SCSL): identifying relevance of a particular column independent of the rest of the schema; 
  \textbf{(Middle + Right)} Table-to-Column Schema Linking (TCSL): first identifying relevant tables then relevant columns.}
  \label{fig:prompts}
  \vspace{-0.95em}
\end{figure}

We explore four different schema linking approaches with a broad spectrum of ability to reduce the false positive rate:
\begin{itemize}[leftmargin=2em]
	\setlength{\itemsep}{2pt}
	\setlength{\parskip}{0pt}
	\setlength{\parsep}{0pt} 
	\item[-] \emph{Single-Column Schema Linking} (SCSL): Model-determined column-wise relevance. The relevance of each column is assessed without context about other columns and tables. 
    The output is a boolean flag per column indicating the relevance of each column. 
    This is considered a more cautious approach less likely to filter out relevant columns. 
    \item[-] \emph{Hybrid SCSL} (HySCSL): SCSL with added keyword matching aiming for higher SLR.
    \item[-] \emph{Table-then-Column Schema Linking} (TCSL): Model-determined table-then-column filtering approach, as proposed in \cite{talaei2024chess} and~\cite{pourreza2023dinsql}. 
    The model first filters the schema to the relevant tables, then filters the columns within those tables. 
    The output is a set of relevant columns and tables. 
    This is a more aggressive filtering approach.
    \item[-] \emph{Hybrid TCSL} (HyTCSL): TCSL with added keyword matching aiming for higher SLR.
\end{itemize}

In our implementation, SCSL and HySCSL use \texttt{GPT-4o-Mini} and TCSL and HyTCSL use \texttt{GPT-4o}. 
We use \texttt{GPT-4o-Mini} with SCSL and HySCSL as \texttt{GPT-4o} shows a negligible gain in our experiments but is much cheaper. 
The prompts used in our implementation are shown in 
\fig{fig:prompts}. 

\tab{tab:snrloss} reports the mean (± stddev) of the false positive rate (FPR) and the schema linking recall (SLR) across our four schema linking techniques and without schema filtering at all (full schema) from $12$ different runs. 
\tab{tab:snrloss} shows that these approaches are robust across runs and that as FPR decreases, \ie more irrelevant columns are filtered, more required columns can be filtered and SLR decreases. 
SLR as obtained from schema linking represents an upper bound on possible EX, where $100\% - SLR$ represents the ratio of queries that will fail in the generation stage due to missing required columns. 

\begin{figure}[t]
  \centering
  \centerline{\includegraphics[width=1.\columnwidth]{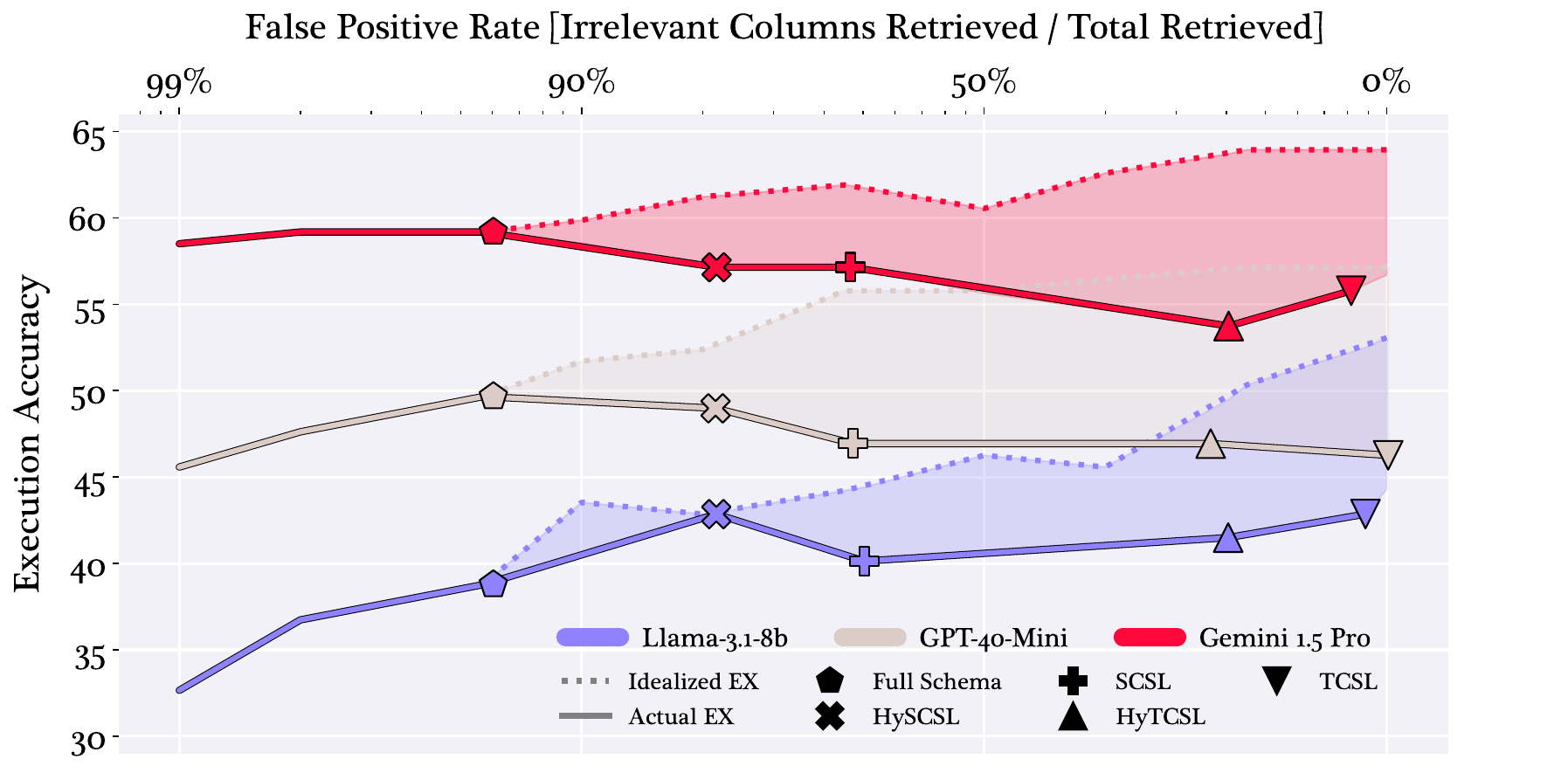}}
  \centerline{\includegraphics[width=0.855\columnwidth]{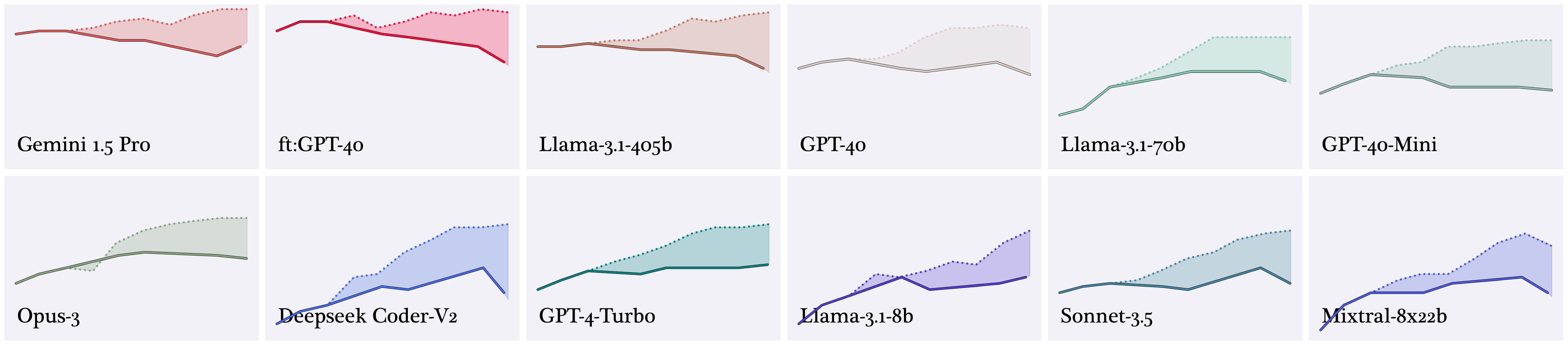}}
  \caption{The execution accuracy (EX) given different schema linking and idealized EX, assuming perfect Schema Linking Recall (SLR)
  as the false positive rate (FPR) varies. 
  EX as a solid line and idealized EX as a dashed line. }
  \label{fig:lossadj}
\end{figure}

We run the same simplified pipeline as our first experiment: 
a retrieval stage containing one of the five schema linking approaches in \tab{tab:snrloss} followed by a zero-shot generation stage. 
We do so across the $12$ selected models (\sect{subsec:models}) and track the associated EX.
\fig{fig:lossadj} shows five EX data points 
given the FPR of the five different schema linking approaches.  
We interpolate between two adjacent (FPR, EX) data points and show EX as a solid line. 
We also add the idealized EX from Experiment 1 as a dashed line. 
The difference between the two lines shows the EX difference due to changes in SLR. 

We observe three different classes of models in our results: models where some schema linking method improves performance (\eg \texttt{Llama 3.1-8b}), models where all schema linking methods degrade performance (\eg \texttt{Gemini 1.5 Pro}), and models where schema linking has only negligible impact on performance (\eg \texttt{GPT-4o-Mini}). 
A model falls into one of these three buckets based on its SQL generation capability. 
More capable models -- e.g. \texttt{Gemini 1.5 Pro}, \texttt{ft:GPT-4o}, \texttt{Llama-3.1-405b} -- end up with a reduction in EX. Less capable models -- e.g. \texttt{Llama 3.1-8b}, \texttt{Mixtral-8x22b}, \texttt{Deepseek Coder-V2} -- end up with a gain in EX.

\emph{Empirical Observation:}
As the model’s SQL generation capability improves, the benefit of schema linking diminishes. In some cases, this can result in a net reduction in accuracy due to missing required columns for generation.

\renewcommand{\arraystretch}{1.3}
\begin{table}[t]
    \centering
    \begin{tabular}{lcc}
        \rowcolor[HTML]{F1F1F7}\hline\hline
        \textbf{Approach} & \textbf{FPR} & \textbf{SLR} \\ \hline\hline
        Without schema filtering (Full Schema) & 94.62 & 100.00 \\
        \rowcolor[gray]{0.97} Hybrid Single-Column Schema Linking (HySCSL) & 82.08 ± 0.44 & 90.36 ± 0.78 \\
        Single-Column Schema Linking (SCSL) & 67.23 ± 0.92 & 88.77 ± 0.75 \\
        \rowcolor[gray]{0.97} Hybrid Table-to-Column Schema Linking (HyTCSL) & 19.85 ± 0.99 & 83.00 ± 0.92 \\
        Table-to-Column Schema Linking (TCSL) & 9.79 ± 0.73 & 77.44 ± 1.34 \\
    \end{tabular}
    \vspace*{3mm}
    \caption{The mean (± stddev) false positive rate (FPR) and schema linking recall  (SLR), across $12$ runs, associated with five schema linking approaches. 
    The approaches are sorted in descending order of their ability to reduce false positives. 
    We report the mean values (± stddev) from $12$ runs.}
    \label{tab:snrloss}
    \vspace{-1em}
\end{table}

\subsection{Experiment 3: Impact of Non-Filtering Stages and Techniques}

\begin{table}[t]
    \centering
    \begin{tabular}{llll}
    \hline\hline
    \rowcolor[HTML]{F1F1F7} & \multicolumn{3}{c}{\textbf{Execution Accuracy (EX)}} \\ \cline{2-4}
    \rowcolor[HTML]{F1F1F7} \multirow{-2}{*}{\textbf{Method}} & \textbf{ft:GPT-4o} & \textbf{Gemini 1.5 Pro} & \textbf{Llama 3.1-405b} \\ \hline\hline
    Full pipeline & 67.35 & 60.54 & 59.18 \\
    \rowcolor[gray]{0.97} w/o Augmentation & 64.63 ($\downarrow$ 2.72) & 60.54 & 59.86 ($\uparrow$ 0.68) \\ 
    w/o Selection & 65.31 ($\downarrow$ 2.04) & 57.82 ($\downarrow$ 2.72) & 58.50 ($\downarrow$ 0.68) \\ 
    \rowcolor[gray]{0.97} w/o Correction & 65.99 ($\downarrow$ 1.36) & 57.14 ($\downarrow$ 3.40) & 55.78 ($\downarrow$ 3.40) \\ 
    w/ TCSL & 62.58 ($\downarrow$ 4.77) & 55.78 ($\downarrow$ 4.76) & 56.46 ($\downarrow$ 2.72) \\ 
    \rowcolor[gray]{0.97} w/ SCSL & 55.78 ($\downarrow$ 11.57) & 55.10 ($\downarrow$ 5.44) & 54.42 ($\downarrow$ 4.76) \\ 
    Base model & 59.18 ($\downarrow$ 8.17) & 57.82 ($\downarrow$ 2.72) & 53.74 ($\downarrow$ 5.44) \\ 
    \end{tabular}
    \label{tab:abl}
    \vspace*{3mm}
    \caption{Ablation of different methods reporting the execution accuracy (EX) for fine-tuned GPT-4o, Gemini 1.5 Pro, and Llama 3.1-405b. The table compares the full pipeline to variations without augmentation, selection, and correction techniques; with Table-to-Column Schema Linking (TCSL) and Single-Column Schema Linking (SCSL); and as base model performance. Reductions ($\downarrow$) or increases ($\uparrow$) in accuracy  compared to the full pipeline are indicated.}
    \label{tab:bird}
    \vspace{-1.9em}
\end{table}

Instead of filtering contextual information through schema linking, we focus on techniques that preserve information. 
We assess the gains of using \emph{augmentation} and \emph{selection} techniques as well as adding a correction stage, which are detailed as follows:\vspace{-0.9em}

\begin{itemize}[leftmargin=2em]
	\setlength{\itemsep}{2pt}
	\setlength{\parskip}{0pt}
	\setlength{\parsep}{0pt} 
	\item[-] \emph{Augmentation:} We add contextual information by: 
    (i) expanding column descriptions and add query hints and 
    (ii) adding structural expectations of the output, \eg expected orderings and aggregations, using CoT planning. 
    \item[-] \emph{Correction:} 
    After generating a candidate SQL query, we iteratively apply corrections through re-generation based on 
    database execution errors~\citep{wang2018robusttexttosqlgenerationexecutionguided}, 
    revision through database administrator instructions~\citep{talaei2024chess}, and 
    model-based feedback similar to Reflexion~\citep{shinn2023reflexionlanguageagentsverbal}. 
    We use these corrections to generate instructions to also augment contextual information. 
    \item[-] \emph{Selection:} We use self-consistency \citep{wang2023selfconsistencyimproveschainthought} to generate multiple responses and select the \emph{most consistent} result. 
    We use selection across the whole pipeline for augmentation, SQL generation, and SQL correction. 
\end{itemize}

%%% TODO (amine):
We implemented a full pipeline using zero-shot generation. 
The pipeline contained augmentation, correction and selection as described above and no schema linking, \ie always providing the full schema. 
We ran an ablation to understand the relative impact of each technique or stage using the three top performing models from Experiment 1: 
\texttt{Llama 3.1-405b}, \texttt{Gemini 1.5 Pro}, and \texttt{ft:GPT-4o}. 
\tab{tab:bird} shows the execution accuracy (EX) of the full pipeline and $6$ other variations: 
without augmentation, correction, or selection, 
with schema linking (TCSL and SCSL from Experiment 2), 
and without any of these techniques \ie providing the full schema alone to a base model. 
We find that all techniques improve accuracy to varying degrees except schema linking with a noticeable difference when evaluating the full pipeline. 

\emph{Empirical Observation:} Each of augmentation, selection, and correction have a noticeable positive impact on generation accuracy. 
Note that even though base models such as \texttt{Gemini 1.5 Pro} may show comparable performance to a fine-tuned \texttt{GPT-4o} in SQL generation, 
they differ when evaluated within end-to-end Text-to-SQL pipelines. 
For instance, augmentation leads to major benefits with \texttt{GPT-4o} when compared with \texttt{Gemini 1.5 Pro}. An interesting research direction is understanding whether the relative benefits from augmentation between models hold across other tasks as well.

\subsection{Discussion and Proposed Approach}
\label{sec:disc}

Our empirical observations reveal several key insights that guide our proposed approach. 

First, as a model's SQL generation capability improves, its ability to retrieve relevant schema elements from a full schema in its input context also improves. 
As such, schema linking for state-of-the-art LLMs is less important if the schema fully fits within the context window. 
However, it is still helpful for models with lower SQL generation accuracy as they struggle with false positives. 
In our approach, \emph{we maximize the use of the LLM context window to minimize filtering required columns}. 

Second, we find that combining augmentation, selection, and correction techniques heavily impacts the accuracy of a Text-to-SQL pipeline. 
However, the impact is not the same when evaluated in an end-to-end fashion even when comparing models with similar generation capability. 
In our approach, \emph{we adopt augmentation, selection, and correction as detailed from the implementation in Experiment 3}. 
We further use \texttt{ft:GPT-4o} as the model of choice for generation since it provides the best end-to-end accuracy and makes the best use of augmentation.  

These design choices yield an approach that ranks first on execution accuracy and second on 
valid efficiency score on the BIRD benchmark. 

\section{Conclusion}
\label{sec:conclusion}
Is it the death of schema linking? For state-of-the-art models, if the schema fits within the context length -- yes. However, for smaller or prior generation models, the accuracy gains from schema linking often justify the potential loss. Additionally, in real-world data-warehousing scenarios, the entire schema often exceeds the context window, requiring a multi-stage information retrieval pipeline. In such cases, we argue for maximally using the LLM context window in picking the top-K relevant columns to retain the necessary schema elements. 
We conclude that while the need for schema linking is highly use-case dependent, its importance is diminishing as costs decrease, context windows widen, and generation capabilities improve.

\section*{Acknowledgements}
We thank Jinyang Li, Yongbin Li and the rest of the BIRD team for evaluating our approach on the BIRD test set. We also thank John Allard for valuable insights regarding fine-tuning.

\clearpage
\bibliographystyle{neurips_2024}
\bibliography{neurips_2024}

\end{document}